\documentclass[conference]{IEEEtran}
\IEEEoverridecommandlockouts
\usepackage{cite}
\usepackage{amsmath,amssymb,amsfonts}
\usepackage{algorithmic}
\usepackage{graphicx}
\usepackage{textcomp}
\usepackage{xcolor}
\usepackage{booktabs}
\usepackage{comment}
\usepackage{url}
\usepackage[utf8]{inputenc}
\def\BibTeX{{\rm B\kern-.05em{\sc i\kern-.025em b}\kern-.08em
    T\kern-.1667em\lower.7ex\hbox{E}\kern-.125emX}}
\begin{document}

\title{Sentiment Analysis on Brazilian Portuguese User Reviews\\
\thanks{This study was financed in part by the Coordenação de Aperfeiçoamento de Pessoal de Nível Superior - Brasil (CAPES) Finance Code 001.}
}

\author{
\IEEEauthorblockN{Frederico Dias Souza}
\IEEEauthorblockA{\textit{Eletrical Engineering Department} \\
\textit{Federal University of Rio de Janeiro}\\
Rio de Janeiro, Brazil \\
fredericods@poli.ufrj.br}
\and
\IEEEauthorblockN{João Baptista de Oliveira e Souza Filho}
\IEEEauthorblockA{\textit{Eletrical Engineering Department} \\
\textit{Federal University of Rio de Janeiro}\\
Rio de Janeiro, Brazil \\
jbfilho@poli.ufrj.br}
}

\maketitle

\begin{abstract}
Sentiment Analysis is one of the most classical and primarily studied natural language processing tasks. This problem had a notable advance with the proposition of more complex and scalable machine learning models. Despite this progress, the Brazilian Portuguese language still disposes only of limited linguistic resources, such as datasets dedicated to sentiment classification, especially when considering the existence of predefined partitions in training, testing, and validation sets that would allow a more fair comparison of different algorithm alternatives. Motivated by these issues, this work analyzes the predictive performance of a range of document embedding strategies, assuming the polarity as the system outcome. This analysis includes five sentiment analysis datasets in Brazilian Portuguese, unified in a single dataset, and  a reference partitioning in training, testing, and validation sets, both made publicly available through a digital repository. A cross-evaluation of dataset-specific models over different contexts is conducted to evaluate their generalization capabilities and the feasibility of adopting a unique model for addressing all scenarios. 

\end{abstract}


\section{Introduction}

Text classification (TC) is a classical natural language processing (NLP) application. The most basic approach for TC consists of extracting specific features from the documents, subsequently feeding them to some classifier responsable for predicting document labels. One of the most popular methods for addressing this feature extraction task is the bag-of-words (BoW). The BoW produces a reduced and simplified representation of an entire document, ignoring aspects like grammar, word appearance order, and semantic relations between words and phrases. A common approach 
is the word frequency, or the term frequency-inverse document frequency (TF-IDF)~\cite{textclassification_survey}. Commonly, the BoW is followed by a classical machine learning (ML) classifier, such as \emph{Logistic Regression}, \emph{Support Vector Machines}, \emph{Gradient Boosting Decision trees}, or \emph{Random Forests}, in sentiment classification tasks. Since most of these models are fast and straightforward to implement and train, they represent a handy baseline. Regardless of their simplicity, such methods can achieve high performance for simple texts, comparable or even better than more complex alternatives. A drawback faced by BoW models is not easily generalize to new tasks and properly deal with the large amounts of training data available nowadays~\cite{dl_textclassification_survey}.

Since the shift in the ML paradigm motivated by the AlexNet \cite{alexnet}, the state-of-the-art models in NLP and Computer Vision mostly include deep learning (DL) architectures \cite{dl_textclassification_survey}. Despite often being more challenging and slower to train, such architectures can easily learn complex patterns and scale to larger datasets. Compared to the classical models, the DL counterparts do not require  hand-crafted feature extraction, leading to automatically learned features during model training~\cite{li2020survey}.

Roughly, a neural-based text classification model can be as simple as a feedforward neural network inputted with a high dimensional vector, defined by some aggregation process of multiple vectors (embeddings) representing the words integrating a document. Popular word embeddings include Word2Vec \cite{word2vec}, GloVe \cite{glove}, and FastText \cite{fasttext}. A remarkable example is due to Iyyer et al. that proposed the Deep Average Network (DAN)  \cite{deep_avg_network}, according to which the embeddings associated with an input sequence of tokens are averaged and the resulting vector is fed through several feedforward layers to produce a vector representing the whole sentence, finally submitted to a simple linear classifier.

Typically, Recurrent Neural Networks (RNNs) can better exploit more complex data patterns than bag-of-words approaches,  accessing more effectively their mutual dependencies, thus better capturing the sentence context~\cite{dl_textclassification_survey}. The most popular variant of the RNNs is the Long Short-Term Memory (LSTM), firstly proposed by Hochreiter and Schmidhuber \cite{first_lstm}, aiming to mitigate the gradient vanishing and exploding problems related to the RNNs \cite{dl_textclassification_survey}. Later, many variants of this model were proposed in the context of text classification, for instance, the remarkable work of Zhou et al.~\cite{bilstm_maxpool}, which proposed a bidirectional LSTM followed by a 2D max-pooling operation.

More recently, the \textit{Transformers}, proposed by Vaswani et al. \cite{attentionallyouneed}, has revolutionized the NLP area. This work has brought a smart alternative to the sequential and slow training faced with the RNNs by solely replacing the recursion mechanism by  multiple attention layers  \cite{dl_textclassification_survey}, whose major distinguishing characteristic refers to the fact that now the training can be performed in parallel. As a result, the complexity of the NLP models could scale largely and huge datasets can be addressed. The possibilities opened by the massive parallel processing available nowadays allowed the Transformers to be applied to the development of large pre\textendash trained models, thus enabling a widely adoption of the transfer learning strategy in NLP applications, as observed years before in the Computer Vision area \cite{cv_nlp_moment}. Among the Transformer based models, the BERT (Bidirectional Encoder Representations from Transformers), proposed by Devlin et al. \cite{bert}, represent one of the most popular architectures. In this direction, the work developed by Sun et al. \cite{bertfinetune} discusses the best practices for fine-tuning BERT for text classification, achieving a remarkable performance. In the opposite direction, i.e., targeting smaller models, the ULMFiT \cite{ulmfit} discusses a light transfer learning model,  achieving the state\textendash of\textendash the\textendash art in text classification.

Despite the recent advances in NLP, the Portuguese language disposes of only limited linguistic resources \cite{cambria_chiong_cornfort}. As Pereira's paper \cite{ptbr_sentimentanalysis_survey} points out, several researchers have been collecting datasets describing a range of text classification problems in Brazilian Portuguese. However, previous works envisaging text classification evaluate a wide variety of datasets or considers different targets to be predicted. Moreover, even in the cases where most of these aspects match, the subsets considered for model evaluation often differ, making a straight comparison of the corresponding results unfair.

Therefore, this work presents a pilot study in which a representative set of sentiment analysis databases in Brazilian Portuguese is systematically organized, made publicity available \footnote{\url{kaggle.com/fredericods/ptbr-sentiment-analysis-datasets}} in \emph{Kaggle}, and analyzed, considering different strategies for text embedding as well as classification techniques. This experimental study also includes a cross\textendash dataset evaluation of the produced models, i.e., one model developed under one dataset is evaluated with another one, and vice and versa. Finally, the performance of a general model trained over a single dataset consolidating all individual ones is conducted.

\section{Experimental study}

\begin{figure*}[t]
\centerline{\includegraphics[width=1.0\textwidth]{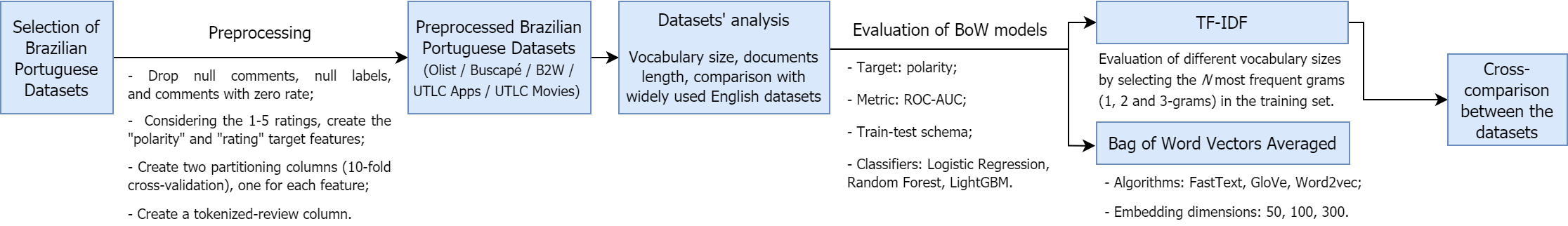}}
\caption{Scheme explaining the work in a summarized way.}
\label{fig:summary_work}
\end{figure*}

The experimental study conducted here included the following phases: dataset collection, pre-processing and analysis; documents embedding, and models evaluation, described in more detail in the following subsections, and illustrated in Figure \ref{fig:summary_work} for convenience. 

\subsection{Dataset collection}

The five representative public user review datasets in Brazilian Portuguese considered in this work are:

\subsubsection{\textbf{Olist}} in 2018, Olist, the largest department store in Brazilian marketplaces, launched the "Brazilian E-Commerce Public Dataset by Olist" \cite{olist_2018} on Kaggle, a database with approximately 100,000 orders from 2016 to 2018 provisioned by several marketplaces in Brazil. Among the different datasets available, this work adopts the \emph{olist order reviews dataset}, which disposes of the user comments plus a label with a satisfaction rate ranging from 1 to 5;

\subsubsection{\textbf{B2W}} in 2019, B2W Digital, one of the most prominent Latin American e-commerce, released the \emph{B2W Reviews01}~\cite{b2w_corpus}, an open corpus of product reviews with more than 130,000 user reviews. This dataset has two target features: the binary label "recommend to a friend" and a user rate from 1 to 5 stars. This work only considered the user rate;

\subsubsection{\textbf{Buscape}} the \emph{Opinando} project \cite{opinando}, proposed by the University of São Paulo (USP) to investigate approaches for concept-level sentiment analysis in Brazilian Portuguese, represents a useful corpora for this work. As described by Hartmann et al. \cite{buscape_corpus}, the \emph{Corpus Buscape} is a large corpus of product reviews in Portuguese, crawled in 2013, integrating  more than 80,000 samples from the \emph{Buscape}, a product and price search website. Unlike the datasets previously described, the Opinando labels' range is from 0 to 5, leading us to remove the comments rated as zero;

\subsubsection{\textbf{UTLCMovies and UTLCApps}} these sets are also part of the \emph{Opinando} project. The \emph{UTLCorpus} \cite{utlc_corpus} is the most extensive set considered here, having more than 2 million reviews. It includes movie reviews collected from the \emph{Filmow}, a famous movie Social network, and mobile apps comments collected from the Google Play Store. Here, the \emph{UTLCorpus} was split into two different datasets: the \emph{UTLCMovies} and the \emph{UTLCApps}. Similar to the Buscape database, reviews with a rating equal to 0 were excluded.

\subsection{Dataset preprocessing and analysis}

We considered two target features for all datasets: polarity and rating (1-5 stars). Polarity is a binary target feature obtained through a label rating process. In this case, a positive review is associated with 4 and 5 stars reviews; a negative, with 1 and 2 stars, while 3 stars occurrences are excluded. The consolidated dataset includes additional columns for indicating data polarity and the partition index assigned to each instance. Such partitions were defined targeting to maintain the original label stratification and for making easy the adoption of both the $k$-fold cross-validation and the classic training-validation-testing hold-out scheme by the practitioner. In this work, the first eight folds were assumed as defining the training set, the ninth fold as the validation set, and the tenth fold as the testing set. 
Besides, we included a column with the user reviews tokenized to facilitate anyone wishing to skip this prepossessing step. 
In this process, we lower-cased all the strings and converted their corresponding letters to the English alphabet, removing the accents and converting "ç" to "c". After that, we obtained tokens with from 2 to 30 alphanumeric characters, disregarding special characters and removing the stop words. Samples with no comments or null labels were also excluded.

Table \ref{tab:dataset_sizes} summarizes the number of classes, samples, and tokens for each dataset. For the sake of comparison, we also included similar numbers for the classic English benchmarks \emph{AG's News}, \emph{DBPedia}, \emph{Yelp Review} (Full and Polarity), and \emph{Amazon Reviews} (Full and Polarity). One may easily note that the datasets in English do not include a validation set and are, in general, larger than the Brazilian counterparts.

\begin{table}[!htbp]
\caption{Number of samples from the Brazilian Portuguese datasets after preprocessing and from some English datasets widely used for text classification.}
\centering
\begin{tabular}{ccccc} 
\toprule
Dataset &
Classes &
\begin{tabular}[c]{@{}c@{}}Train\\samples\end{tabular} &
\begin{tabular}[c]{@{}c@{}}Validation\\samples\end{tabular} &
\begin{tabular}[c]{@{}c@{}}Test\\samples\end{tabular}\\
\midrule
Olist (Polarity)        & 2 & 30 k   & 4 k   & 4 k \\
Olist (Rating)          & 5 & 33 k   & 4 k   & 4 k \\
B2W (Polarity)          & 2 & 93 k   & 12 k  & 12 k \\
B2W (Rating)            & 5 & 106 k  & 13 k  & 13 k \\
Buscape (Polarity)      & 2 & 59 k   & 7 k   & 7 k \\
Buscape (Rating)        & 5 & 68 k   & 8 k   & 8 k \\
UTLCApps (Polarity)    & 2 & 775 k  & 97 k  & 97 k \\
UTLCApps (Rating)      & 5 & 832 k  & 104 k & 104 k \\
UTLCMovies (Polarity)  & 2 & 952 k  & 119 k & 119 k \\
UTLCMovies (Rating)    & 5 & 1190 k & 149 k & 149 k \\
All combined (Polarity) & 2 & 1909 k & 239 k & 239 k \\
All combined (Rating)   & 5 & 2229 k & 279 k & 279 k \\
\midrule
AG’s News         & 4  & 120 k   & - & 7.6 k  \\
DBPedia           & 14 & 560 k   & - & 70 k  \\
Yelp (Polarity)   & 2  & 560 k   & - & 38 k \\
Yelp (Rating)     & 5  & 650 k   & - & 50 k \\
Yahoo! Answers    & 10 & 1400 k & - & 60 k \\
Amazon (Polarity) & 2  & 3600 k & - & 400 k\\
Amazon (Rating)   & 5  & 3000 k & - & 650 k\\
\bottomrule
\end{tabular}\label{tab:dataset_sizes}
\end{table}

Table \ref{tab:ptbr_dataset_vocab} compares the document and vocabulary sizes for each one of the datasets, accounting for only the tokens/grams occurrences that appeared more than five times in the corpus. As expected, increasing the dataset size leads to a more extensive vocabulary, except for the Buscape, smaller than B2W but with longer sentences. Overall, the English datasets have richer content than the Portuguese ones, with considerably longer sentences. This fact indicates that the Portuguese datasets may be less benefited from more complex models than English.

\begin{table}[!htbp]
\caption{Document length (number of tokens) and vocabulary size from Brazilian Portuguese and English datasets.}
\centering
\begin{tabular}{cccccc} 
\toprule
Dataset &
\begin{tabular}[c]{@{}c@{}}Mean\\length\end{tabular} &
\begin{tabular}[c]{@{}c@{}}Median\\length\end{tabular} &
\begin{tabular}[c]{@{}c@{}}Vocab size\\(1 gram)\end{tabular} &
\begin{tabular}[c]{@{}c@{}}Vocab size\\(1 and 2 grams)\end{tabular}\\
\midrule
Olist        & 7  & 6  & 3.272  & 8.491 \\
Buscape      & 25 & 17 & 13.470 & 52.769\\
B2W          & 14 & 10 & 12.758 & 47.929\\
UTLCApps     & 7  & 5  & 28.283 & 179.227\\
UTLCMovies   & 21 & 10 & 69.711 & 635.869\\
All combined & 15 & 7  & 86.234 & 884.398\\
\midrule
AG’s News & 21 & 20 & 24.713 & 96.070\\
DBPedia & 30 & 30 & 110.755 & 521.403\\
Yelp Reviews & 68 & 50 & 70.494 & 1.303.148\\
Yahoo! Answers & 11 & 3 & 65.534 & 464.409\\
Amazon Reviews & 38 & 33 & 176.464 & 3.475.911\\
\bottomrule
\end{tabular}\label{tab:ptbr_dataset_vocab}
\end{table}

In addition, to better infer the level of similarity among different datasets, Table \ref{tab:words_in_common} shows the percentage of words/tokens shared between them. Most of the words included in the Olist, the smallest and disposing of shorter sentences, is present in other datasets, on average 93.5\%. In turn, the much larger UTLCApps and UTLCMovies detain the highest percentage of words (an average of 69.6\% and 84.4\%, respectively) common to all other datasets. This fact might have particularity contributed to the experimental finding that their  models generalized quite well to other datasets, as will be seeing in the following.

\begin{table}[!htbp]
\caption{Percentage of words in common between datasets. Example: 89.5\% of the Olist vocabulary is contained in the Buscape vocabulary and 21.7\% of the Buscape vocabulary is in the Olist vocabulary. Rows and columns average (avg) do not count the 100\% from the diagonal entries.}
\centering
\begin{tabular}{ccccccc} 
\toprule
Dataset &
Olist &
Buscape &
B2W &
\begin{tabular}[c]{@{}c@{}}UTLC\\Apps\end{tabular} &
\begin{tabular}[c]{@{}c@{}}UTLC\\Movies\end{tabular} &
Avg\\
\midrule
Olist       & 100.0 & 21.7  & 25.3  & 10.6  & 4.4   & 15.5\\
Buscape     & 89.5  & 100.0 & 73.5  & 37.1  & 16.3  & 54.1\\
B2W         & 98.5  & 69.6  & 100.0 & 35.6  & 15.8  & 54.9\\
UTLCApps   & 91.8  & 77.8  & 79.0  & 100.0 & 29.6  & 69.6\\
UTLCMovies & 94.1  & 84.4  & 86.1  & 72.9  & 100.0 & 84.4\\
Average     & 93.5  & 63.4  & 66.0  & 39.1  & 16.5  & -\\
\bottomrule
\end{tabular}\label{tab:words_in_common}
\end{table}

Finally, Table \ref{tab:label_distribution} analyzes the experimental distribution of the target features: polarity and rating in these datasets. Despite the higher frequency of the positive label, the distribution varies a lot, which may hinder the performance of adopting a single model for all datasets.

\begin{table}[!htbp]
\caption{Labels distribution for each dataset. In the upper part, we present the 5-point scale rate target feature, In the lower, the polarity target feature, where 0 represents reviews with negative reviews (1 and 2 points) and 1 corresponds to the positive reviews (4 and 5 points).}
\centering
\begin{tabular}{ccccccc} 
\toprule
Label &
Olist &
Buscape &
B2W &
\begin{tabular}[c]{@{}c@{}}UTLC\\Apps\end{tabular} &
\begin{tabular}[c]{@{}c@{}}UTLC\\Movies\end{tabular} &
All\\
\midrule
1 & 22.0 & 3.7  & 20.7 & 16.4 & 2.4  & 8.9\\
2 & 5.3  & 4.3  & 6.3  & 4.5  & 6.8  & 5.8\\
3 & 8.8  & 13.4 & 12.3 & 6.8  & 20.0 & 14.4\\
4 & 14.5 & 39.5 & 24.4 & 11.4 & 36.7 & 26.4\\
5 & 49.4 & 39.1 & 36.2 & 60.8 & 34.0 & 44.5\\
\midrule
0 (1-2) & 30.0 & 9.2 & 30.8 & 22.5 & 11.6  & 17.2\\
1 (3-4) & 70.0 & 90.8 & 69.2 & 77.5 & 88.4 & 82.8\\
\bottomrule
\end{tabular}\label{tab:label_distribution}
\end{table}

The rationale behind making this consolidated dataset publicity available is providing a valuable resource for those interested in the sentiment analysis of commercial textual data in Portuguese.

\subsection{Documents' embeddings}

The first embedding approach considered here was the \textit{tf-idf} (term-frequency inverse-document frequency), introduced by Jones \cite{first_tfidf}. Described by Equation \ref{eq:tfidf}, the weight $w_{i,j}$ assigned to the word $i$ in the document $j$ corresponds to the multiplication of term-frequency $tf_{i,j}$ (number of occurrences of $i$ in $j$) by the inverse document frequency $idf_{i}$, defined as the logarithm of the ratio of the total number of documents $N$ to $df_{i}$, the latter equal to the number of documents containing the word $i$ \cite{later_tfidf}. Note that the term frequency rewards words with high occurrence in a specific document, and the inverse document frequency penalizes frequent words  in a document collection. 

\begin{equation} \label{eq:tfidf}
\left\{\begin{matrix}
w_{i,j} = tf_{i,j}.idf_{i}
\\ 
idf_{i} = log(\frac{N}{df_i})
\end{matrix}\right.
\end{equation}

Our analysis considered a vocabulary consisted of 1 to 3-grams, i.e., defined by individual words (1-grams), pairs (2-grams),  and triplets (3-grams) of successive words. We evaluated different vocabulary sizes by selecting the $N$ most frequent n-grams, limited to a maximum of 500.000. Besides, only the n-grams appearing more than five times in the corpus were considered. Below, we summarize the vocabulary sizes considered for each dataset.

\begin{itemize}
\item \textbf{Olist}: $50$, $100$, $300$, $1000$, $5000$, $9574$;
\item \textbf{Buscape}: $50$, $100$, $300$, $1000$, $5000$, $10000$, $25000$, $50000$, $61294$;
\item \textbf{B2W}: $50$, $100$, $300$, $1000$, $5000$, $10000$, $25000$, $50000$, $55936$;
\item \textbf{UTLCApps}: $50$, $100$, $300$, $1000$, $5000$, $10000$, $25000$, $50000$, $75000$, $100000$, $213311$;
\item \textbf{UTLCMovies}: $50$, $100$, $300$, $1000$, $5000$, $10000$, $25000$, $50000$, $75000$, $100000$, $250000$, $500000$;
\item \textbf{All combined}: $50$, $100$, $300$, $1000$, $5000$, $10000$, $25000$, $50000$, $75000$, $100000$, $250000$, $500000$.
\end{itemize}

As an alternative to \textit{tf-idf}, we considered representing each document by the average value of all embedding vectors of the words integrating it. The word embeddings evaluated in this case were \emph{Word2Vec} \cite{word2vec}, \emph{GloVe} \cite{glove}, and \emph{FastText} \cite{fasttext}, all
previously trained with  Brazilian Portuguese corpora and released by the NILC Word Embeddings Repository~\cite{nilc_repo}. In the process of identifying which embedding vector would be assigned to each word, the words were first lowercased, converted to the English alphabet, i.e., had their accents removed, and the occurrences of "ç" were replaced by "c".
The words not found in the embedding dictionary were ignored. The experiments assumed embedding sizes equal to 50, 100, and 300. The computation of the document embedding only considered words appearing more than five times in the corpus.

\subsection{Classification algorithms}

The classification techniques considered in this study include \emph{Logistic Regression} (LR), \emph{Random Forest} (RF) \cite{randomforest}, and \emph{LightGBM} (LGBM) \cite{lgbm}. The first is a standard classification model. The other two represent ensemble approaches based on classification trees, exploring bagging and boosting strategies, respectively, thus disposing of more learning capabilities and higher computational burden. The choice of the \emph{LightGBM} rather than the standard gradient boosting was motivated by its lower computational cost.

The experimental analysis also included the evaluation of the generalization capacity of models trained over different datasets. In this case, a model trained with one dataset was tested with instances from another dataset and vice-versa. A single model trained with all datasets integrated was also produced. This analysis only considered the \textit{tf-idf} embedding scheme.

\section{Results and Discussion}

For the following models, for the sake of simplicity, we adopted the hyperparameters defined in their corresponding packages as standard. Additionally, training and validation sets were concatenated during models development. In all tests, the area under the ROC curve (ROC-AUC) was the figure of merit.

\subsection{Average document embedding strategy}

First, we analyzed sentiment analysis models based on the average document  embeddings strategy. Among all experiments involving different techniques, sizes, and datasets, the best ROC-AUC for the testing sets were achieved by the \emph{LightGBM}, which represents the most powerful classifier among those evaluated. Table \ref{tab:tfidf_results} summarizes the values obtained. Only the results related to the best combinations of embeddings techniques, embedding sizes, datasets, and classifiers are reported. Overall, there is a significant increase in the predictive performance with the growth of the embedding size. Also, FastText outperforms GloVe, which supersedes Word2Vec. The 300-dimensional FastText embedding performs best for all datasets. Thus, the embedding technique and the dimensionality of the embedding vector have shown a profound impact on the model performance.

\begin{table}[!htbp]
\caption{ROC-AUC (\%) for the the bag of word vectors averaged embeddings, followed by LightGBM, where "FastText 300" refers to 300-dimensional FastText word embeddings. The best result for each dataset is underlined.}
\centering
\begin{tabular}{ccccccc} 
\toprule
Model &
Olist &
Buscape &
B2W &
\begin{tabular}[c]{@{}c@{}}UTLC\\Apps\end{tabular} &
\begin{tabular}[c]{@{}c@{}}UTLC\\Movies\end{tabular} &
All\\
\midrule
FastText 50 & 93.9 & 86.2 & 94.3 & 91.5 & 79.3 & 86.5\\
FastText 100 & 94.9 & 87.4 & 95.2 & 92.3 & 81.7 & 87.8\\
FastText 300 & \underline{95.5} & \underline{88.5} & \underline{96.1} & \underline{93.2} & \underline{83.9} & \underline{89.0}\\
GloVe 50 & 93.2 & 85.2 & 93.0 & 90.8 & 78.0 & 85.6\\
GloVe 100 & 93.8 & 86.5 & 94.4 & 91.8 & 80.3 & 86.9\\
GloVe 300 & 95.1 & 87.7 & 95.6 & 92.6 & 83.1 & 88.2\\
Word2Vec 50 & 92.8 & 83.9 & 92.4 & 89.7 & 75.4 & 83.9\\
Word2Vec 100 & 94.0 & 85.6 & 93.6 & 91.1 & 78.3 & 85.7\\
Word2Vec 300 & 94.8 & 87.8 & 95.1 & 92.4 & 82.0 & 87.6\\
\bottomrule
\end{tabular}\label{tab:tfidf_results}
\end{table}

\begin{figure*}[!htbp]
\centerline{\includegraphics[width=1.0\textwidth]{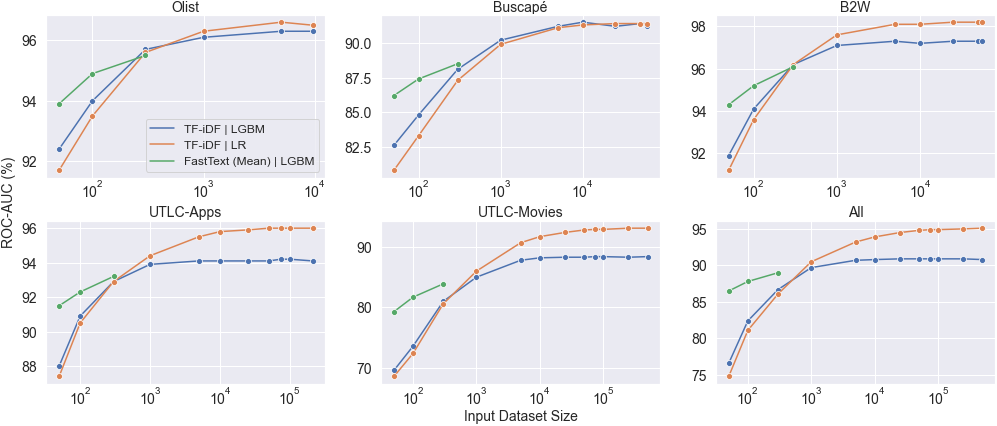}}
\caption{ROC-AUC (\%) per Input Dataset Size}
\label{fig:tfidf_bow_results}
\end{figure*}

\subsection{Tf-idf}

Second, the study focused on the \textit{tf-idf} embedding schema. Here, the results restrict to the \emph{Logistic Regression} and \emph{LightGBM}, the classification techniques that achieved the best results in this analysis. Figure \ref{fig:tfidf_bow_results} depicts the values of ROC-AUC as a function of the number of n-grams for each dataset. The increase in the input data dimensionality has brought a considerable predictive gain, regardless of the dataset considered, for the range of dimensions from 50 to 5,000. The gains, however, level off for dimensions above 10,000. 

Also, in general, both classifiers exhibited a similar performance for small vocabulary sizes, with a minor advantage for the \emph{LightGBM}. For higher dimensions, the \emph{Logistic Regression} outperforms the latter, with a moderate advantage, except to the \emph{Buscape} to which a tie is observed.

The plot includes the results of 50, 100, and 300\textendash dimensional FastText word embeddings. For smaller dimensions, 50 and 100, the document embedding schema is significantly better than \textit{tf-idf}. For dimension 300, both performed similarly.

Our findings seem to be in coherence with Zhang et al.~\cite{zhang_cnn}, which reported a superior performance of the \textit{tf-idf}. In our case, this result may be related to the simplicity of the aggregation process, according to which meaningful and irrelevant words similarly contribute to the resulting document embedding, which might have led to a significant information loss. Therefore, for the scope of the evaluation considered here, a good  trade-off between performance and complexity for the 
\textit{tf-idf} strategy is achieved with 10,000 dictionary words.

\subsection{Models cross-comparison} 

The cross-model evaluation considered the \textit{tf-idf} scheme followed by the \emph{Logistic Regression}. This analysis consisted of evaluating each model trained in one dataset using data from another one. Additionally, a model referred to as "All Combined" was produced for all datasets concatenated. For each dataset, we considered the maximum number of n-grams possible, limited to 500,000 grams, to take full advantage of the common words shared between the datasets.

Table \ref{tab:cross_results} shows the results of these experiments. In all cases, the best performances were attained with the datasets over which the models were trained upon, as expected. However, the model produced with all datasets combined does not perform much worse, detaining the second-best result, with an absolute difference over the first ranging from 0.1\% to 0.5\%. As a result, one may conclude that a single model could generalize well to all datasets. In this analysis, the higher drop observed with the UTLC Movies is related to the fact that it is the unique dataset not covering product reviews, as pointed in Table \ref{tab:words_in_common}. Thus, its vocabulary tends to be the most specific among those considered here. Furthermore, the Buscape model, which is associated with the second smallest dataset, has also faced difficulties when generalizing to other datasets, a behavior that might be related to its longer sentences (Table \ref{tab:ptbr_dataset_vocab}).

\begin{table}[!htbp]
\caption{ROC-AUC (\%) for each training and evaluation dataset combination. For example: when the model is trained with Buscape and evaluated with Olist, the Test ROC-AUC is 91.6. When the opposite occurs, it is 85.3. "Delta" refers to the difference in performance between the model trained on all datasets and the model trained on the specific dataset.}
\centering
\begin{tabular}{ccccccc} 
\toprule
Model &
Olist &
Buscape &
B2W &
\begin{tabular}[c]{@{}c@{}}UTLC\\Apps\end{tabular} &
\begin{tabular}[c]{@{}c@{}}UTLC\\Movies\end{tabular} &
All\\
\midrule
Olist        & \underline{96.5} & 85.3 & 96.5 & 90.3 & 66.7 & 82.8\\
Buscape      & 91.6 & \underline{91.4} & 96.2 & 90.9 & 73.1 & 85.5\\
B2W          & 96.1 & 88.4 & \underline{98.2} & 92.4 & 73.9 & 87.4\\
UTLCApps    & 94.5 & 88.7 & 96.4 & \underline{96.0} & 76.9 & \underline{90.2}\\
UTLCMovies  & 79.0 & 80.9 & 86.4 & 85.6 & \underline{93.1} & 88.9\\
All combined & \underline{96.4} & \underline{91.1} & \underline{98.0} & \underline{95.9} & \underline{92.6} & \underline{95.1}\\
\midrule
Delta & -0.1 & -0.3 & -0.2 & -0.1 & -0.5 & -\\
\bottomrule
\end{tabular}\label{tab:cross_results}
\end{table}

\section*{Conclusion}
In this work, we unified five public annotated datasets of user reviews in Brazilian Portuguese   targeting sentiment classification. We observed that these datasets are considerably poorer (i.e., they detain smaller vocabularies and shorter sentences) than their English counterparts. We proposed pre-defined partitions for each target (polarity and 1-5 rating), hoping to encourage other interested researchers in evaluating the performance of alternative models in the same partitions, facilitating   further analysis as well as direct comparisons with the results presented here.

We also evaluated different document embedding strategies. The models employing average word embeddings performed better with the FastText and higher dimensional vectors. However, such models were outperformed by those adopting the \textit{tf-idf} with vocabulary sizes equal to or greater than 1,000 words. Regarding the study relative to the models' generalization among different datasets, one expected finding was that a single model trained with all datasets performed worse than models specifically trained over specific datasets, despite the difference observed was  small. Thus, for this problem, it is possible to provide a good single model.

In future works, we expect to evaluate if more complex machine learning algorithms and text representation strategies could even improve the results achieved with  these simple datasets. In addition, since these datasets are constituted by texts freely inserted by web users, we intend to study the cost-effectiveness of additional preprocessing schemes to better deal with words and phrases written in a more informal fashion.


\section*{References}
\bibliographystyle{IEEEtran}
\renewcommand{\refname}{}
\bibliography{paper}

\end{document}